\pdfoutput=1
\documentclass[11pt]{article}

\usepackage[]{naacl2021}
\usepackage[english]{babel}
\usepackage[utf8]{inputenc}
\usepackage{subfigure}
\usepackage{caption}
\usepackage{tabularx}
\usepackage{multicol}
\usepackage[nottoc]{tocbibind}
\usepackage{times}
\usepackage{latexsym}
\usepackage{graphicx}

\usepackage{chngcntr}
\counterwithin{figure}{section}
\counterwithin{table}{section}

\usepackage[T1]{fontenc}
\usepackage[utf8]{inputenc}

\usepackage{microtype}

\title{Leveraging recent advances in Pre-Trained
Language Models for Eye-Tracking Prediction}

\author{Varun Madhavan \\
  IIT Kharagpur \\ 
    {\tt varun.m.iitkgp}\\
  {\tt @gmail.com} \\\And
  
  Aditya Girish Pawate \\
  IIT Kharagpur \\
  {\tt adityagirish}\\
  {\tt pawate@gmail.com} \\\And
  
    Shraman Pal \\
  IIT Kharagpur \\
    {\tt shramanpal12345}\\
  {\tt @gmail.com} \\\And
  
    Abhranil Chandra \\
  IIT Kharagpur \\
     {\tt abhranil.chandra}\\
  {\tt @gmail.com} \\
  }

\usepackage{array}
\usepackage[font=small]{caption}
\usepackage{multicol, blindtext}
\newcolumntype{L}{>{\centering\arraybackslash}m{0.35\textwidth}}
\newcolumntype{D}{>{\centering\arraybackslash}m{0.65\textwidth}}
\newcolumntype{E}{>{\centering\arraybackslash}m{0.09\textwidth}}

\begin{document}
\maketitle

\begin{abstract}
Cognitively inspired Natural Language Processing uses human-derived behavioral data like eye-tracking data, which reflect the semantic representations of language in the human brain to augment the neural nets to solve a range of tasks spanning syntax and semantics with the aim of teaching machines about language processing mechanisms. In this paper, we use the ZuCo 1.0 and ZuCo 2.0 dataset containing the eye-gaze features to explore different linguistic models to directly predict these gaze features for each word with respect to its sentence. We tried different neural network models with the words as inputs to predict the targets. And after lots of experimentation and feature  engineering finally devised a novel architecture consisting of RoBERTa Token Classifier with a dense layer on top for language modeling and a stand-alone model consisting of dense layers followed by a transformer layer for the extra features we engineered. Finally, we took the mean of the outputs of both these models to make the final predictions. We evaluated the models using mean absolute error (MAE) and the R2 score for each target.
\end{abstract}

\section{Introduction}
When reading, we humans process language “automatically” without reflecting on each step — we string words together into sentences, understand the meaning of spoken and written ideas, and process language without thinking too much about how the underlying cognitive process happens. This process generates cognitive signals that could potentially facilitate natural language processing tasks. Our understanding of the field of language processing is highly dependent on accurately modeling eye-tracking features (example of cognitive data) as they provide millisecond-accurate records of where humans look while reading, thus shedding some light on human attention mechanisms and language comprehension skills. Eye movement data is used extensively in various fields including but not limited to Natural Language Processing and Computer Vision. The recent introduction of standardized datasets has enabled the comparison of the capabilities of different cognitive modeling and linguistically motivated approaches to model and analyze human patterns of reading.%\newblock

\section{Problem Description}

\begin{table}[ht]
    \centering
    \begin{tabular}{|p{0.25\linewidth} | p{0.6\linewidth}|}
    \hline
    \textbf{Feature Name} & \textbf{Feature Description}\\
    \hline
    \textbf{nFix} & Number of fixations on the word\\
    \hline
    \textbf{FFD} & Duration of the first fixation\\
    \hline
    \textbf{TRT} & Total fixation duration, including regressions\\
    \hline
    \textbf{GPT} & Sum of all prior fixations\\
    \hline
    \textbf{fixProp} & Proportion of participants that fixated the current word \\
    \hline
    \end{tabular}
    \caption{Target Variables }
    \label{table:targets}
\end{table}
Our brain processes language and generates cognitive processing data such as gaze patterns and brain activity. These signals can be recorded while reading. The problem is formulated as a regression task to predict the five eye-tracking features for each token of a sentence in the test set. Each training sample consists of a token in a sentence and the corresponding features. We are required to predict the five features: nFix, FFD, TRT, GPT, and fixProp. We were given a CSV file with sentences to train our model. Tokens in the sentences were split in the same manner as they were presented to the participants during the reading experiments. Hence, this did not necessarily follow linguistically correct tokenization. Sentence endings were marked with an <EOS> symbol. A prediction had to be made for each token for all the five target variables. The evaluation metric used was the mean absolute error.

\section{Dataset}

We use the eye-tracking data of the Zurich Cognitive Language Processing Corpus (ZuCo 1.0 and ZuCo 2.0) \cite{hollenstein2020zuco} recorded during normal reading. The training and test data contains 800 and 191 sentences respectively.In recent years, collecting these signals has become increasingly easy and less expensive \cite{DBLP:conf/ijcai/PapoutsakiSLD0H16}; as a result, using cognitive features to improve NLP tasks has become more popular. The data contains features averaged over all readers and scaled in the range of 0 to 100, facilitating evaluation via mean absolute average (MAE).
The 5 eye-tracking features included in the dataset are shown in \ref{table:targets}. The heatmap of the correlation matrix of the 5 target variables have been shown in Figure 1. It is observed that apart from GPT all the features are correlated with each other to a large extent.

\begin{figure}
    \centering
    \includegraphics[width=0.35\textwidth]{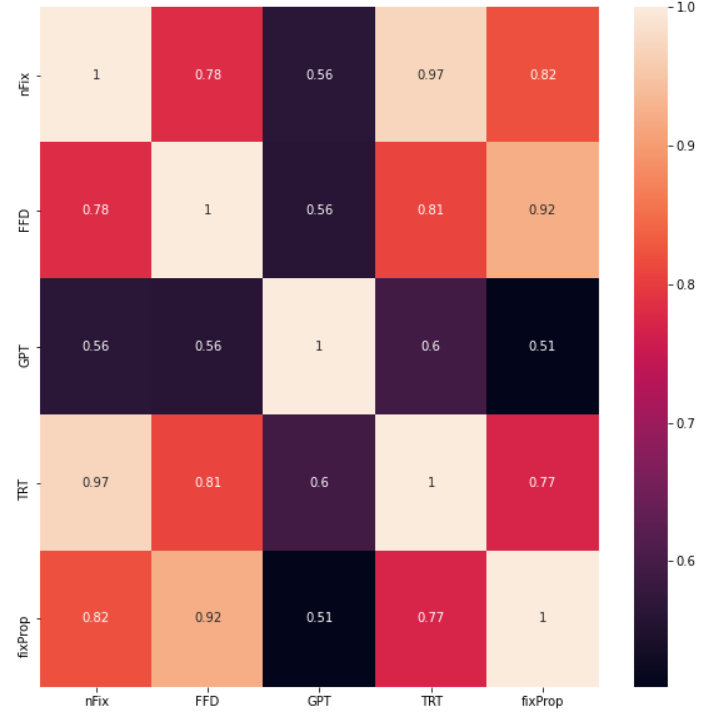}
    \caption{Correlation Heatmap of Targets}
    \label{fig:my_label}
\end{figure}

\section{Preprocessing}
We first removed the <EOS> token at the end of each sentence. We looked at similar word \cite{CLIFTON2007341}, \cite{DEMBERG2008193} and \cite{Just80atheory} to find features that we could add. As GPT consistently performed the worst in all our initial experiments, we tried replacing GPT with a new feature (TRT - GPT) during training. The results improved slightly and we decided to keep this change for the remaining experiments. To incorporate word level characteristics (i.e characteristics that belong to a word; eg: word length) and syntactic information we introduced the following features -
\begin{itemize}
    \item Number of characters in a word ($word len$)
    \item Number of characters in the lemmatized word subtracted from the number of characters in the word. Obtained using NLTK package \cite{bird-loper-2004-nltk} ($lem word len$)
    \item If a word was a stopword or not, we appended 1 if stopword and -1 otherwise($stopword$)
    \item If a word was a number, we appended 1 if it was a number and -1 otherwise($number$)
    \item If a word was the at end of the sentence or not, we appended 1 if last word and -1 otherwise ($endword$). Our hypothesis for adding the endword was that it had a strong correlation with GPT as endwords had significantly higher GPT values compared to the other words.
    \item Part of Speech Tags for each word in a sentence ($pos tag$). We one hot encoded the POS TAGS which gave us a sparse matrix of 20 features. The POS tags have a high statistical correlation with all the target variables as they represent the semantic structure of a sentence. Obtained using NLTK package.
    \item TF-IDF value of each word by considering each sentence as a separate document and each word as a separate term($tfidf$). For the tfidf score, words unrecognized by the in-built tokenizer due to being single digit numbers, having “-” etc were replaced by 0.01 if it was among “a”, “A” or “I” and 0.1 otherwise. This provides a hyperparameter(TF-IDF Error) that could be tuned for better results. Obtained using sklearn package. \cite{scikit-learn}
\end{itemize} 

Scatter plots of word length(number of characters) and the 5 target variables have been shown. It is observed that there is quite a positive correlation between them. we expected the last word of the sentence to have an impact on the GPT score which is evident from the barplot as shown in figure 2f. The GPT score for a endword is usually quite high compared to a non endword.
We also created barplots of stopword and the target variables. It is evident that for stopwords the values of all the targets variables are comparitively lower which points to stopword being a significant feature.
The effect of adding these features is further explored as an ablation study in section 6.3. 
\begin{figure}[h]
  \begin{minipage}[b]{0.22\textwidth}
    \centering
    \includegraphics[width=\textwidth]{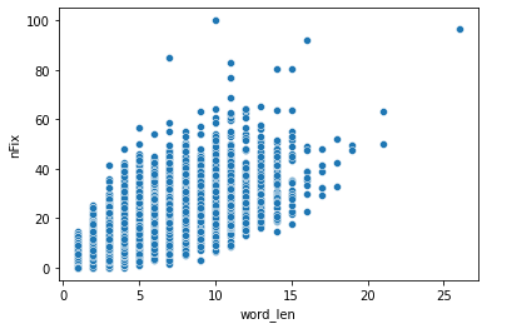}
    \caption{nFix vs \newline wordlen}
    \label{fig:2a}
  \end{minipage}%
  \begin{minipage}[b]{0.22\textwidth}
    \centering
    \includegraphics[width=\textwidth]{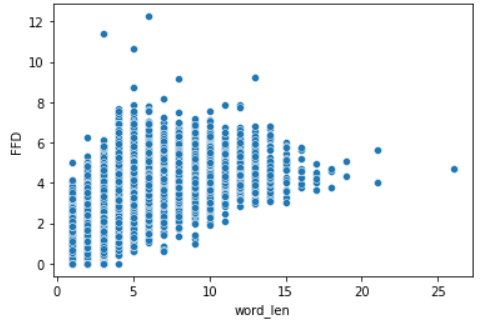}
    \caption{FFD vs \newline wordlen}
    \label{fig 2b}
  \end{minipage}
\end{figure}

\begin{figure}[h]
  \begin{minipage}[b]{0.22\textwidth}
    \centering
    \includegraphics[width=\textwidth]{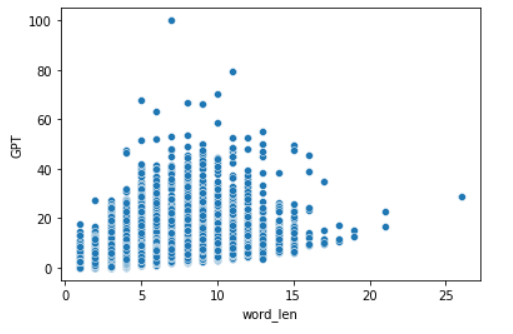}
    \caption{GPT vs \newline wordlen}
    \label{fig:2c}
  \end{minipage}%
  \begin{minipage}[b]{0.22\textwidth}
    \centering
    \includegraphics[width=\textwidth]{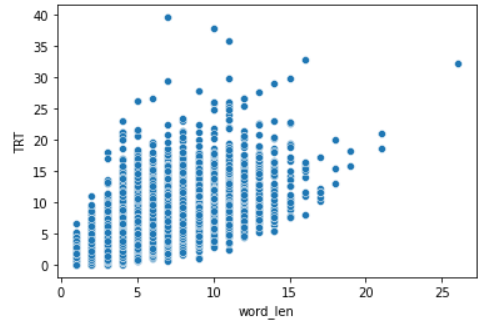}
    \caption{TRT vs \newline wordlen}
    \label{fig:2d}
  \end{minipage}
\end{figure}

\begin{figure}[h]
  \begin{minipage}[b]{0.22\textwidth}
    \centering
    \includegraphics[width=\textwidth]{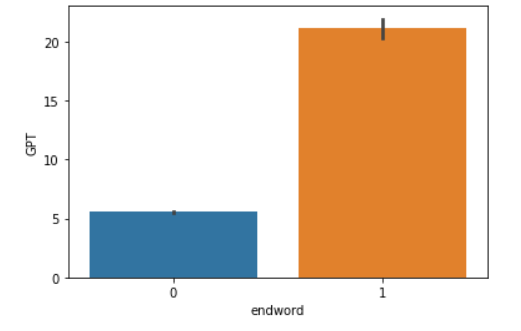}
    \caption{GPT vs \newline endword}
    \label{fig:2f}
  \end{minipage}
  \begin{minipage}[b]{0.22\textwidth}
    \centering
    \includegraphics[width=\textwidth]{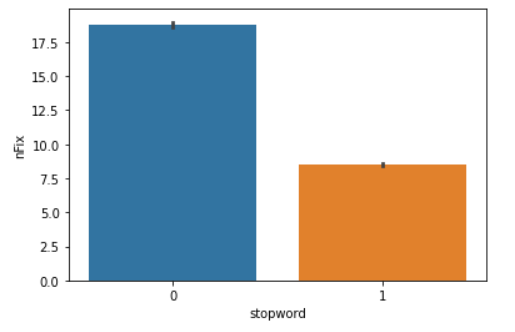}
    \caption{nFix vs \newline stopword}
    \label{fig:2g}
  \end{minipage}%
\end{figure}

\begin{figure}[h]
  \begin{minipage}[b]{0.22\textwidth}
    \centering
    \includegraphics[width=\textwidth]{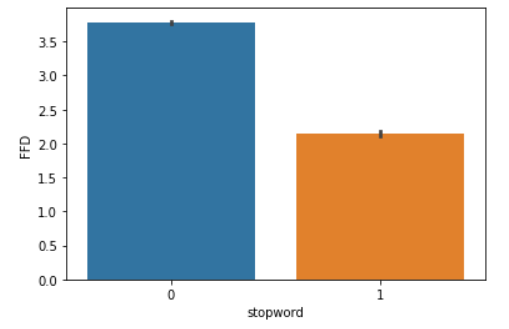}
    \caption{FFD vs \newline stopword}
    \label{fig:2h}
  \end{minipage}
  \begin{minipage}[b]{0.22\textwidth}
    \centering
    \includegraphics[width=\textwidth]{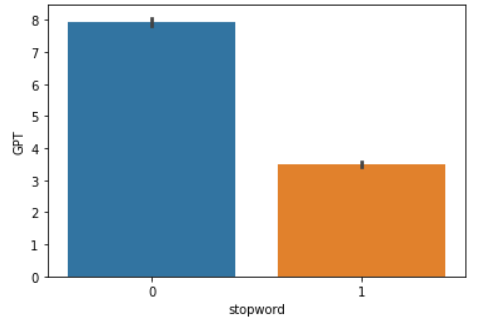}
    \caption{GPT vs \newline stopword}
    \label{fig:2i}
  \end{minipage}%
\end{figure}

\begin{figure}[b]
  \begin{minipage}[b]{0.22\textwidth}
    \centering
    \includegraphics[width=\textwidth]{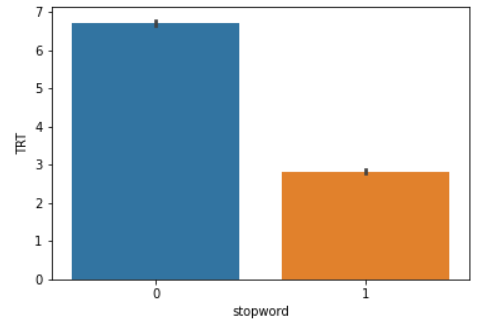}
    \caption{TRT vs \newline stopword}
    \label{fig:2j}
  \end{minipage}%
  \begin{minipage}[b]{0.22\textwidth}
    \centering
    \includegraphics[width=\textwidth]{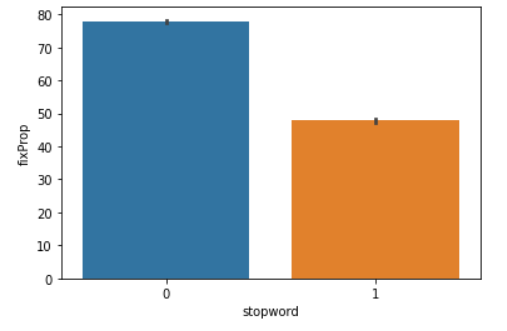}
    \caption{fixProp vs \newline stopword}
    \label{fig:2k}
  \end{minipage}%
\end{figure}

\section{Model Description}
The model has two major components, i.e. the \textbf{Feature Model} and the \textbf{Language Model}, which encode the features and the sentence respectively. The resulting encoded representations are mean pooled and passed to a fully connected layer and sigmoid activation to output a (5x1) score vector for each token in the sentence. The score vector represents the values of nFix, FFD, TRT, GPT and fixProp respectively.   

\begin{figure}[h]
  \begin{minipage}[b]{0.23\textwidth}
    \centering
    \includegraphics[width=\textwidth]{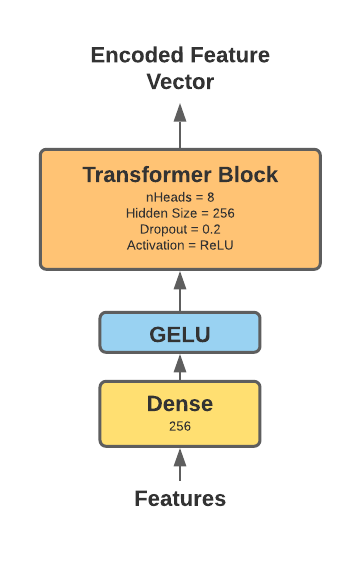}
    \caption{Feature Model}
    \label{fig:1}
  \end{minipage}%
  \begin{minipage}[b]{0.23\textwidth}
    \centering
    \includegraphics[width=\textwidth]{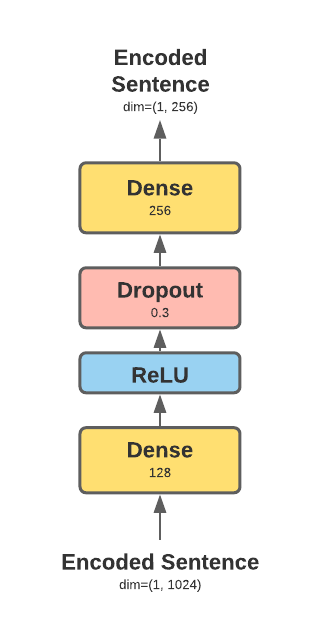}
    \caption{Language Head}
    \label{fig:2}
  \end{minipage}
\end{figure}

\begin{figure*}[t]
    \centering
    \includegraphics[height=4cm]{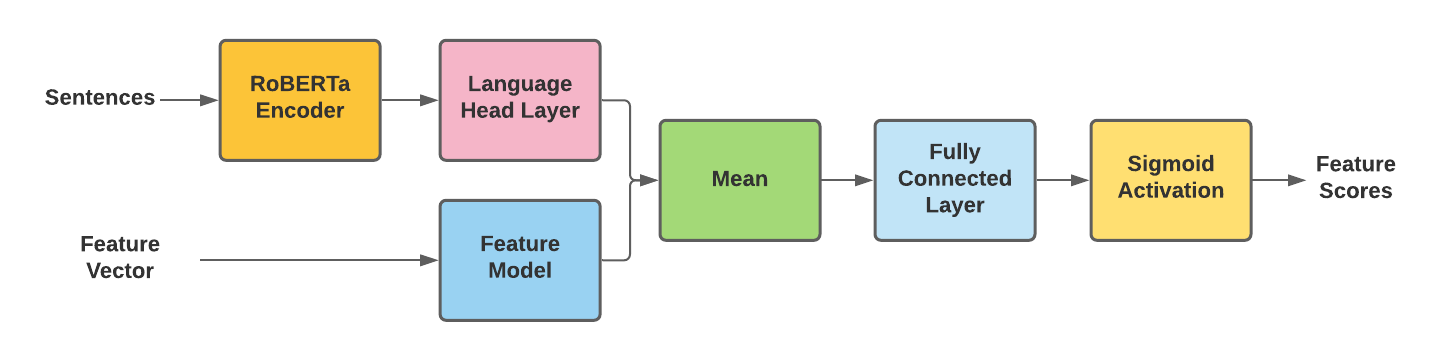}
    \caption{Complete Model Flow}
    \label{fig:my_label}
\end{figure*}

\section{Experiments}

\subsection{Evaluation Metric}
We utilised the R2 score metric instead of the Mean Absolute Error score because R2 provides a better understanding of the model's performance in absence of a baseline score. As the R2 score has an upper bound of 1 it is possible to understand how good the model is performing for a particular value of R2 score. 

\[R^2(y, \hat{y}) = 1 - \frac{\sum_{i=1}^{n} (y_i - \hat{y}_i)^2}{\sum_{i=1}^{n} (y_i - \bar{y})^2}\]

\subsection{Models} We experimented with a variety of models, in particular recent advances in attention-based\cite{vaswani2017attention} pre-trained language models like RoBERTa \cite{DBLP:journals/corr/abs-1907-11692}, ELECTRA \cite{ELECTRA}, BERT \cite{devlin2019bert}, DistilBERT\cite{sanh2020distilbert}. Each language model was used to encode text as described in section 5 and the rest of the pipeline was kept the same. We compare this method with \cite{hollenstein2019advancing}, which employed GloVe(200d) \cite{pennington2014glove} and a Bidirectional LSTM \cite{hochreiter1997long} to encode sentences. In addition to GloVe, we also tried the more recent ConceptNet Numberbatch embeddings (300D) \cite{speer2018conceptnet} that have been shown to outperform GloVe on a variety of tasks. The results for all experiments are tabulated in Table \ref{results}.

\subsection{Features}
In addition to various architectures, we tried a number of feature engineering techniques with inferred/transformed features as described in section 4. We performed an ablation study with each feature to judge the importance/benefit of each adding each of the new features. 

We note that increasing the number of features does not always help, so analysing the data we saw that most of the POS tags rarely occurred and there were 6-7 frequently occurring tags. We decided to club the remaining significant tags under their respective larger categories (eg: noun phrases(NNP) were clubbed into noun category(NN)) and marked the remaining small number of categories as <UNK> while also one-hot encoding all of them. Lastly we standardized the targets for better training stability and it was applied after transforming GPT to TRT-GPT.

\begin{table}[h!]
    \centering
    \begin{tabular}{|>{\centering\arraybackslash}m{0.15\textwidth}|>{\centering\arraybackslash}m{0.15\textwidth}|>{\centering\arraybackslash}m{0.1\textwidth}|}
    \hline
        \textbf{Feature Name} & \textbf{Feature Description} & \textbf{Average Score and Std Deviation}  \\
        \hline
         \textbf{word len} & Number of characters in the word & 0.6145 \newline 0.07363\\
         \hline
         \textbf{is stopword} & Whether a word is a stopword & 0.6475  0.06043\\
         \hline
         \textbf{is number} & Whether a word is a number & 0.6194  0.06731\\
         \hline
         \textbf{lem word len} & Difference in the number of characters in the word and its lemmatized form & 0.6125  0.06649\\
         \hline
         \textbf{is endword} & Whether the word is the last word of its sentence & 0.5918  0.10606\\
         \hline
         \textbf{tfidf score} & TF-IDF score of the word in the corresponding sentence & 0.6265  0.06367\\
         \hline
         \textbf{pos tag} & Part of Speech Tag of the word & 0.6284 0.059786 \\
      \hline 
    \end{tabular}
    \caption{Features and their Description and the Mean and Std Deviation of R2 scores of target variables}
    \label{tab:my_label}
\end{table}

\begin{table}[h!]
    \centering
    \begin{tabular}{|c|c|}
        \hline
         \textbf{Hyperparameter} & \textbf{Value} \\
         \hline
         \textbf{Batch Size} & 4 \\
         \hline
         \textbf{Num Warm Up Steps} & 4\\
         \hline
         \textbf{Learning Rate} & 3e-5 \\
         \hline
         \textbf{Max epochs} & 120\\
         \hline
         \textbf{Beta 1} & 0.91 \\ 
         \hline
         \textbf{Delta} & 0.0001\\
         \hline
         \textbf{Beta 2} & 0.998 \\
         \hline
         \textbf{Early stopping patience} & 8\\
         \hline
         \textbf{Weight Decay} & 1e-5 \\
         \hline
         \textbf{TF-IDF Error} & 0.1\\
         \hline
         \textbf{Validation Split Ratio} & 0.2 \\
         \hline
         \textbf{Training Split Ratio} & 0.8 \\
         \hline
    \end{tabular}
    \caption{Hyperparameter Values}
    \label{tab:my_label}
\end{table}

\begin{itemize}
  \item \textbf{Effect of scaling: } Due to a large number of features with a large difference in their means, it is helpful to scale the features to the same range/same mean and standard deviation. Hence we compared two scaling methods - 
    \begin{itemize}
        \item \textbf{Min-max scaling: } \newline $X_{norm} = (X - X_{min})/(X_{max} - X_{min})$  
        \item \textbf{Standard scaling: } \newline $X_{norm} = (X - X_{mean})/(X_{std})$  
    \end{itemize}

\begin{table*}[h]
\centering
\begin{tabular}{|>{\centering\arraybackslash}m{0.11\textwidth}|c|c|*{5}{>{\centering\arraybackslash}m{0.06\textwidth}|} }
\hline
\textbf{Model} & \textbf{Feature Norm} & \textbf{Extra Features} & \textbf{nFix} & \textbf{FFD} & \textbf{GPT} & \textbf{TRT} & \textbf{fixProp}\\
\hline
\textbf{RoBERTa} & Std Scl & STT & 0.8842 &	0.9246 &	0.7343 &	0.8823 &	0.9509 \\
\hline
\textbf{RoBERTa} & Min Max & All &	0.8835 &	0.9132 &	0.7383 &	0.8542 &	0.9461\\
\hline
\textbf{RoBERTa} & Min Max & STT &	0.8417&	0.9090&	0.6456&	0.8152&	0.9492\\
\hline
\textbf{BERT} & Min Max & All & 0.7433 &	0.8528&	0.4574&	0.6925&	0.9173\\
\hline
\textbf{BiLSTM Minibatch} & Min Max & All + GloVE & 0.6595 &	0.7275&	0.6281&	0.5988&	0.7583\\
\hline
\textbf{BiLSTM Minibatch} & Min Max & All + NB\footnote{https://github.com/commonsense/conceptnet-numberbatch} & 0.6379 &	0.7203 & 0.5992&	0.5388&	0.7592\\
\hline
\textbf{RoBERTa Token} & Std Scl & All & 0.7038 &	0.6512&	0.4697&	0.6877&	0.7146\\
\hline
\textbf{BERT Token} & Std Scl & All & 0.7231&	0.7107&	0.3440&	0.6376&	0.7635\\
\hline
\textbf{BERT} & Std Scl & All &	0.6497&	0.6456&	0.5409&	0.6088&	0.7171\\
\hline
\textbf{BiLSTM MiniBatch} & Std Scl & All+GloVE &	0.6850&	0.5976&	0.5267&	0.5940&	0.7158\\
\hline
\textbf{RoBERTa Token} & Std Scl & All & 0.6483	&0.5861	&0.5475	&0.6280&	0.6566\\
\hline
\textbf{RoBERTa} & Std Scl & All & 0.6360&	0.5869	&0.4798	&0.6061&	0.6623\\
\hline
\textbf{BiLSTM Stochastic} & Std Scl & All+GloVE & 0.6282&	0.6247&	0.4394	&0.5860&	0.6893\\
\hline
\textbf{RoBERTa Token} & Std Scl & All &	0.6349&	0.5914&	0.4767&	0.6123&	0.6452\\
\hline
\textbf{RoBERTa Token} & Std Scl & All & 0.5991&	0.5847&	0.5417&	0.5686&	0.6524\\
\hline
\textbf{BERT Token} & Std Scl & All & 0.6182 & 0.5677	&0.4495	&0.6026&	0.6960\\
\hline
\textbf{BERT} & Std Scl & All & 0.6019&	0.5847&	0.4159	&0.5771&	0.6404
\\
\hline
\textbf{BERT} & Std Scl & All & 0.5681	&0.5437	&0.4891&	0.5170&	0.6371\\
\hline
\textbf{DistilBert} & No Scaling & All & 0.4804&	0.5292&	0.0983&	0.4183&	0.6324\\
\hline
\end{tabular}
\caption{Compiled results of all the top performing models. STT means Single Target Training (one model to predict one feature). Numberbatch refers to the ConceptNet Numberbatch word embeddings used \cite{speer2017conceptnet}. The architecture was kept same (except for input dimensions), the learning rate was reduced for smaller number of features to achieve optimal values which we observed through trial and error.}\label{results}
\end{table*}  

With the same model (multi-target with RoBERTa encoder), using standard scaling gave an R2 score of 0.62 while min-max scaling gave a score of 0.81. The reason for min-max scaling giving significantly better performance might be because the targets in this dataset are always positive. 
  
\item \textbf{Single Target vs. Multi-Target models: } With a large number of input features and a relatively small datasets, we hypothesized that large pre-trained models may fail to converge to a suitable minima in the limited fine-tuning. Hence, we trained separate instances of the same model on each of nFix, FFD, TRT, GPT and fixProp. This further increased the R2 score on the validation set to 0.88.

\end{itemize}

%\columnbreak
\section{Results}

In \ref{results}, we tabulate the R2 scores of each of our models on the validation set.

%\resizebox{1\linewidth}{3cm}{%
%\resizebox{\linewidth}{3cm}{

\clearpage

\bibliographystyle{acl_natbib}
\bibliography{main}

\begin{thebibliography}{17}
\expandafter\ifx\csname natexlab\endcsname\relax\def\natexlab#1{#1}\fi

\bibitem[{Bird and Loper(2004)}]{bird-loper-2004-nltk}
Steven Bird and Edward Loper. 2004.
\newblock \href {https://www.aclweb.org/anthology/P04-3031} {{NLTK}: The
  natural language toolkit}.
\newblock In \emph{Proceedings of the {ACL} Interactive Poster and
  Demonstration Sessions}, pages 214--217, Barcelona, Spain. Association for
  Computational Linguistics.

\bibitem[{Clark et~al.(2020)Clark, Luong, Le, and Manning}]{ELECTRA}
Kevin Clark, Minh-Thang Luong, Quoc~V. Le, and Christopher~D. Manning. 2020.
\newblock \href {http://arxiv.org/abs/2003.10555} {Electra: Pre-training text
  encoders as discriminators rather than generators}.

\bibitem[{Clifton et~al.(2007)Clifton, Staub, and Rayner}]{CLIFTON2007341}
Charles Clifton, Adrian Staub, and Keith Rayner. 2007.
\newblock \href
  {https://doi.org/https://doi.org/10.1016/B978-008044980-7/50017-3} {Chapter
  15 - eye movements in reading words and sentences}.
\newblock In Roger~P.G. {Van Gompel}, Martin~H. Fischer, Wayne~S. Murray, and
  Robin~L. Hill, editors, \emph{Eye Movements}, pages 341--371. Elsevier,
  Oxford.

\bibitem[{Demberg and Keller(2008)}]{DEMBERG2008193}
Vera Demberg and Frank Keller. 2008.
\newblock \href
  {https://doi.org/https://doi.org/10.1016/j.cognition.2008.07.008} {Data from
  eye-tracking corpora as evidence for theories of syntactic processing
  complexity}.
\newblock \emph{Cognition}, 109(2):193--210.

\bibitem[{Devlin et~al.(2019)Devlin, Chang, Lee, and
  Toutanova}]{devlin2019bert}
Jacob Devlin, Ming-Wei Chang, Kenton Lee, and Kristina Toutanova. 2019.
\newblock \href {http://arxiv.org/abs/1810.04805} {Bert: Pre-training of deep
  bidirectional transformers for language understanding}.

\bibitem[{Hochreiter and Schmidhuber(1997)}]{hochreiter1997long}
Sepp Hochreiter and J{\"u}rgen Schmidhuber. 1997.
\newblock Long short-term memory.
\newblock \emph{Neural computation}, 9(8):1735--1780.

\bibitem[{Hollenstein et~al.(2019)Hollenstein, Barrett, Troendle, Bigiolli,
  Langer, and Zhang}]{hollenstein2019advancing}
Nora Hollenstein, Maria Barrett, Marius Troendle, Francesco Bigiolli, Nicolas
  Langer, and Ce~Zhang. 2019.
\newblock \href {http://arxiv.org/abs/1904.02682} {Advancing nlp with cognitive
  language processing signals}.

\bibitem[{Hollenstein et~al.(2020)Hollenstein, Troendle, Zhang, and
  Langer}]{hollenstein2020zuco}
Nora Hollenstein, Marius Troendle, Ce~Zhang, and Nicolas Langer. 2020.
\newblock \href {http://arxiv.org/abs/1912.00903} {Zuco 2.0: A dataset of
  physiological recordings during natural reading and annotation}.

\bibitem[{Just and Carpenter(1980)}]{Just80atheory}
Marcel~Adam Just and Patricia~A. Carpenter. 1980.
\newblock A theory of reading: from eye fixations to comprehension.
\newblock \emph{PSYCHOLOGICAL REVIEW}, 87(4).

\bibitem[{Liu et~al.(2019)Liu, Ott, Goyal, Du, Joshi, Chen, Levy, Lewis,
  Zettlemoyer, and Stoyanov}]{DBLP:journals/corr/abs-1907-11692}
Yinhan Liu, Myle Ott, Naman Goyal, Jingfei Du, Mandar Joshi, Danqi Chen, Omer
  Levy, Mike Lewis, Luke Zettlemoyer, and Veselin Stoyanov. 2019.
\newblock \href {http://arxiv.org/abs/1907.11692} {Roberta: {A} robustly
  optimized {BERT} pretraining approach}.
\newblock \emph{CoRR}, abs/1907.11692.

\bibitem[{Papoutsaki et~al.(2016)Papoutsaki, Sangkloy, Laskey, Daskalova,
  Huang, and Hays}]{DBLP:conf/ijcai/PapoutsakiSLD0H16}
Alexandra Papoutsaki, Patsorn Sangkloy, James Laskey, Nediyana Daskalova, Jeff
  Huang, and James Hays. 2016.
\newblock \href {http://www.ijcai.org/Abstract/16/540} {Webgazer: Scalable
  webcam eye tracking using user interactions}.
\newblock In \emph{Proceedings of the Twenty-Fifth International Joint
  Conference on Artificial Intelligence, {IJCAI} 2016, New York, NY, USA, 9-15
  July 2016}, pages 3839--3845. {IJCAI/AAAI} Press.

\bibitem[{Pedregosa et~al.(2011)Pedregosa, Varoquaux, Gramfort, Michel,
  Thirion, Grisel, Blondel, Prettenhofer, Weiss, Dubourg, Vanderplas, Passos,
  Cournapeau, Brucher, Perrot, and Duchesnay}]{scikit-learn}
F.~Pedregosa, G.~Varoquaux, A.~Gramfort, V.~Michel, B.~Thirion, O.~Grisel,
  M.~Blondel, P.~Prettenhofer, R.~Weiss, V.~Dubourg, J.~Vanderplas, A.~Passos,
  D.~Cournapeau, M.~Brucher, M.~Perrot, and E.~Duchesnay. 2011.
\newblock Scikit-learn: Machine learning in {P}ython.
\newblock \emph{Journal of Machine Learning Research}, 12:2825--2830.

\bibitem[{Pennington et~al.(2014)Pennington, Socher, and
  Manning}]{pennington2014glove}
Jeffrey Pennington, Richard Socher, and Christopher~D. Manning. 2014.
\newblock \href {http://www.aclweb.org/anthology/D14-1162} {Glove: Global
  vectors for word representation}.
\newblock In \emph{Empirical Methods in Natural Language Processing (EMNLP)},
  pages 1532--1543.

\bibitem[{Sanh et~al.(2020)Sanh, Debut, Chaumond, and
  Wolf}]{sanh2020distilbert}
Victor Sanh, Lysandre Debut, Julien Chaumond, and Thomas Wolf. 2020.
\newblock \href {http://arxiv.org/abs/1910.01108} {Distilbert, a distilled
  version of bert: smaller, faster, cheaper and lighter}.

\bibitem[{Speer et~al.(2017)Speer, Chin, and Havasi}]{speer2017conceptnet}
Robyn Speer, Joshua Chin, and Catherine Havasi. 2017.
\newblock \href {http://aaai.org/ocs/index.php/AAAI/AAAI17/paper/view/14972}
  {{ConceptNet} 5.5: An open multilingual graph of general knowledge}.
\newblock pages 4444--4451.

\bibitem[{Speer et~al.(2018)Speer, Chin, and Havasi}]{speer2018conceptnet}
Robyn Speer, Joshua Chin, and Catherine Havasi. 2018.
\newblock \href {http://arxiv.org/abs/1612.03975} {Conceptnet 5.5: An open
  multilingual graph of general knowledge}.

\bibitem[{Vaswani et~al.(2017)Vaswani, Shazeer, Parmar, Uszkoreit, Jones,
  Gomez, Kaiser, and Polosukhin}]{vaswani2017attention}
Ashish Vaswani, Noam Shazeer, Niki Parmar, Jakob Uszkoreit, Llion Jones,
  Aidan~N. Gomez, Lukasz Kaiser, and Illia Polosukhin. 2017.
\newblock \href {http://arxiv.org/abs/1706.03762} {Attention is all you need}.

\end{thebibliography}

\end{document}